\documentclass[letterpaper]{article} 
\usepackage{aaai24}  
\usepackage{times}  
\usepackage{helvet}  
\usepackage{courier}  
\usepackage[hyphens]{url}  
\usepackage{graphicx} 
\usepackage{natbib}  
\usepackage{caption} 
\usepackage{times}
\usepackage{epsfig}
\usepackage{graphicx}
\usepackage{amsmath}
\usepackage{amssymb}

\usepackage{subcaption}
\usepackage{graphicx}
\usepackage{caption}
\usepackage{booktabs}
\usepackage{bm}
\usepackage{multirow}
\usepackage{bbding}
\usepackage{color}
\usepackage{times}
\usepackage{epsfig}
\usepackage{graphicx}
\usepackage{amsmath}
\usepackage{amssymb}

\usepackage{subcaption}
\usepackage{booktabs}
\usepackage{bm}
\usepackage{multirow}
\usepackage{bbding}
\usepackage[linesnumbered,ruled,vlined]{algorithm2e}

\title{Learning Image Demoir{\'e}ing from Unpaired Real Data}
\author{
    Yunshan Zhong$^{1,2}$, Yuyao Zhou$^{2,3}$, Yuxin Zhang$^{2,3}$, Fei Chao$^{2,3}$, Rongrong Ji$^{1,2,3,4}$\thanks{Corresponding Author}
}
\affiliations{
    $^1$Institute of Artificial Intelligence, Xiamen University. \\ $^2$Key Laboratory of Multimedia Trusted Perception and Efficient Computing, \\
    Ministry of Education of China, Xiamen University. \\
    $^3$Department of Artificial Intelligence, School of Informatics, Xiamen University.\\$^4$Peng Cheng Laboratory.
}

\usepackage{bibentry}

\usepackage[switch]{lineno}

\begin{document}

\maketitle

\begin{abstract}

This paper focuses on addressing the issue of image demoir{\'e}ing. Unlike the large volume of existing studies that rely on learning from paired real data, we attempt to learn a demoir{\'e}ing model from unpaired real data, i.e., moir{\'e} images associated with irrelevant clean images. 
The proposed method, referred to as Unpaired Demoir{\'e}ing (UnDeM), synthesizes pseudo moir{\'e} images from unpaired datasets, generating pairs with clean images for training demoir{\'e}ing models.
To achieve this, we divide real moir{\'e} images into patches and group them in compliance with their moir{\'e} complexity.
We introduce a novel moir{\'e} generation framework to synthesize moir{\'e} images with diverse moir{\'e} features, resembling real moir{\'e} patches, and details akin to real moir{\'e}-free images. 
Additionally, we introduce an adaptive denoise method to eliminate the low-quality pseudo moir{\'e} images that adversely impact the learning of demoir{\'e}ing models.
We conduct extensive experiments on the commonly-used FHDMi and UHDM datasets.
Results manifest that our UnDeM performs better than existing methods when using existing demoir{\'e}ing models such as MBCNN and ESDNet-L. Code: \url{https://github.com/zysxmu/UnDeM}.

\end{abstract}

\section{Introduction}
\label{sec:intro}

Contemporary society is awash with electronic screens for presenting images, text, video, \emph{etc}. With the widespread availability of portable camera devices such as smartphones, people have grown accustomed to using them for quick information recording. 
Unfortunately, a common issue arises from the intrinsic interference between the camera’s color filter array (CFA) and LCD subpixel layout of the screen~\cite{yu2022towards}, resulting in captured pictures being contaminated with some rainbow-shape stripes, which are also known as moir{\'e} patterns~\cite{sun2018moire,yang2017demoireing}. 
These moir{\'e} patterns involve varying thickness, frequencies, layouts, and colors, which degrade the perceptual quality of captured pictures. Consequently, there has been considerable academic and industrial interest in developing demoir{\'e}ing algorithms to rectify the issue.

\begin{figure}[t]
\centering
\includegraphics[width=0.75\linewidth]{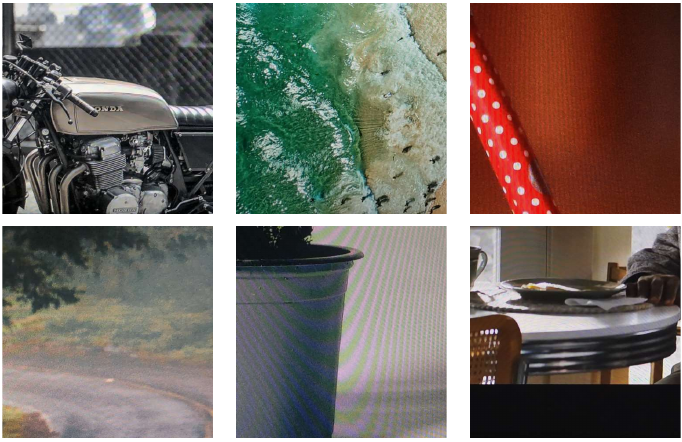}
\caption{Illustration of image moir{\'e}. Natural moir{\'e} patterns are complex in varying thicknesses, frequencies, layouts, and colors across images and within an image.}
\label{fig:moire-pattern}
\end{figure}

Primitive research on demoir{\'e}ing are mostly established upon image priors~\cite{dabov2007image,cho2011image} or traditional machine learning methods~\cite{liu2015moire,yang2017textured}, which are demonstrated to be inadequate for tackling moir{\'e} patterns of drastic variations~\cite{zheng2021learning}. Fortunately, the fashionable convolutional neural networks (CNNs) have become a de facto infrastructure for the success of various computer vision tasks including the image demoir{\'e}ing~\cite{he2020fhde,cheng2019multi,he2019mop,liu2020wavelet,sun2018moire,yuan2019aim,zheng2021learning,yu2022towards,liu2018demoir,gao2019moire}. 
These CNN-based methods are typically trained on extensive pairs of moir{\'e}-free and moir{\'e} images in a supervised manner to model the demoir{\'e}ing mapping.
However, it is challenging to collect paired images given the fact in Fig.\,\ref{fig:moire-pattern} that natural moir{\'e} patterns are featured with varying thicknesses, frequencies, layouts, and colors~\cite{zheng2021learning}.
We can easily access to the moir{\'e} images as well as moir{\'e}-free images, but they are mostly unpaired.
Although many studies try to capture image pairs from digital screens~\cite{he2020fhde,yu2022towards}, their quality is barricaded by three limitations.
First, acquiring high-quality image pairs involves professional camera position adjustments and even special hardware~\cite{yu2022towards}.
Second, burdensome manpower is required to select well-aligned moir{\'e}-free and moir{\'e} pairs.
Third, the captured moir{\'e} contents are very unitary under highly-controlled lab environments.
However, image pairs full of more diverse moir{\'e} patterns are more expected for improving demoir{\'e}ing models.

%

%
%
%
%
%
%

\begin{figure*}[ht]
    \centering
    \begin{subfigure}[b]{\textwidth}
        \centering
        \includegraphics[width=0.85\linewidth]{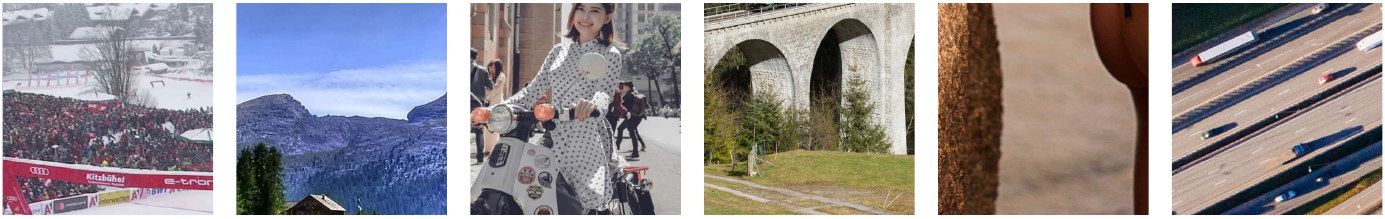} 
        \caption{Moir{\'e}-free images.}
        \label{fig:real-clear}
    \end{subfigure}
    \\
    \begin{subfigure}[b]{\textwidth}
        \centering
        \includegraphics[width=0.85\linewidth]{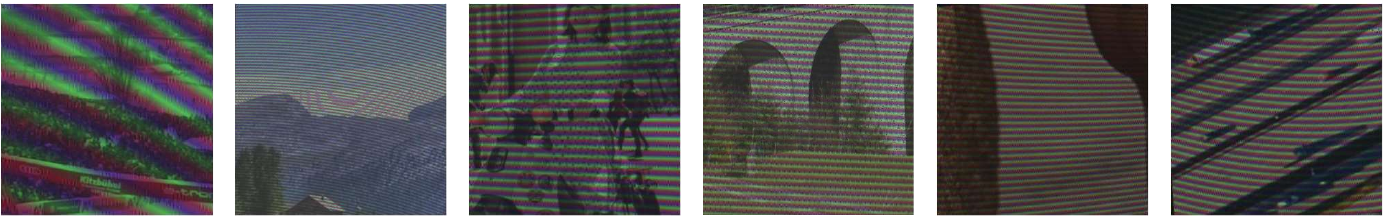} 
        \caption{Pseudo moir{\'e} images by shooting simulation.}
        \label{fig:shooting}
    \end{subfigure}
    \\
    \begin{subfigure}[b]{\textwidth}
        \centering
        \includegraphics[width=0.85\linewidth]{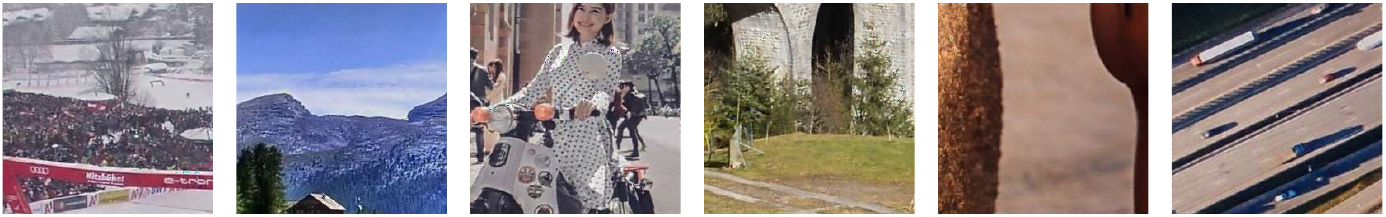} 
        \caption{Pseudo moir{\'e} images by cyclic learning.}
        \label{fig:baseline}
    \end{subfigure}
    \\
    \begin{subfigure}[b]{\textwidth}
        \centering
        \includegraphics[width=0.85\linewidth]{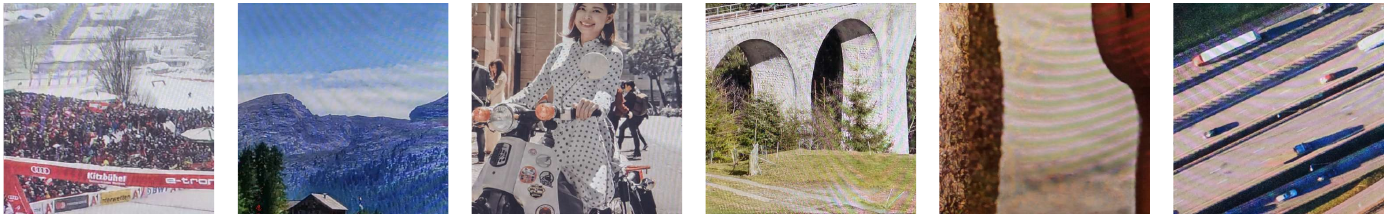} 
        \caption{Pseudo moir{\'e} images by our UnDeM.}
        \label{fig:our-mo}
    \end{subfigure}
\caption{Visual examples of (a) real moir{\'e}-free images; (b) pseudo moir{\'e} images by shooting simulation~\cite{niu2021morie}; (c) pseudo moir{\'e} images by cyclic learning~\cite{park2022unpaired}; (d) pseudo moir{\'e} images by our UnDeM. Compared with details-missing and inauthentic pseudo moir{\'e} images by shooting simulation and \cite{park2022unpaired}, ours result in more diverse moir{\'e} patterns and preserve more details of the moir{\'e}-free images. Best view by zooming in.}
\label{fig:insight}
\end{figure*}


Synthesizing moir{\'e} images has therefore attracted increasing attention recently. Given Fig.\,\ref{fig:real-clear} illustrated moir{\'e}-free screenshots, shooting simulation methods~\cite{liu2018demoir,yuan2019aim,niu2021morie} simulate the aliasing between CFA and screen’s LCD subpixel to produce corresponding paired moir{\'e} images in Fig.\,\ref{fig:shooting}. 
However, the synthetic images are insufficient to capture characteristics of real moir{\'e} patterns, leading to a large domain gap as we analyze from two aspects.
First, the synthetic moir{\'e} images are much darker and cannot well seize the light quality, destroying the context of the viewing environment and obscuring the image details.
Second, the synthetic moir{\'e} patterns lack authenticity, as the thicknesses, frequencies, layouts, and colors of moir{\'e} stripes are almost the same within an image.
In Table\,\ref{tab:fhdmi} and Table\,\ref{tab:uhdm} of the experimental section, we apply shooting simulated moir{\'e} images to train demoir{\'e}ing CNNs. Our results manifest that the trained models generalize poorly to the natural-world test datasets. 
In ~\cite{park2022unpaired}, Park \emph{et al}. introduced a cyclic moir{\'e} learning method and we observe a better performance than shooting simulation in Table\,\ref{tab:fhdmi} and Table\,\ref{tab:uhdm}. However, the generated pseudo moir{\'e} fails to accurately model real moir{\'e} patterns as illustrated in Fig.\,\ref{fig:baseline}, leading to limited performance.
Therefore, it is desired to develop a better method to improve the synthetic moir{\'e} images.

In this paper, we present one novel method, dubbed as UnDeM, to learn demoir{\'e}ing from unpaired real moir{\'e} and clean images that are fairly easy to collect, for example, by performing random screenshots and taking random photos from the digital screen. 
As displayed in Fig.\,\ref{fig:our-mo}, the basic objective of our UnDeM is to synthesize moir{\'e} images that possess mori{\'e} features as the real moir{\'e} images and details as the real moir{\'e}-free images.
The synthesized pseudo moir{\'e} images then form pairs with the real moir{\'e}-free images for training demoir{\'e}ing networks.
To this end, as shown in Fig.\,\ref{Preprocessing}, we first split images into patches. These moir{\'e} patches are further grouped using a moir{\'e} prior that takes into consideration frequency and color information in each patch~\cite{zhang2023real}. Consequently, moir{\'e} patches within each group fall into similar complexity such that they can be better processed by the individual moir{\'e} synthesis network.
Specifically, the introduced synthesis network contains four modules including a moir{\'e} feature encoder to extract moir{\'e} features of real moir{\'e} patches, a generator to synthesize pseudo moir{\'e} patches, a discriminator to identify real or pseudo moir{\'e} patches, and a content encoder to retain content information of real clean patches in synthesized pseudo moir{\'e} patches.
The whole framework is conducted in an adversarial training manner~\cite{goodfellow2014generative} for a better moir{\'e} image generation.

Before paired with real moir{\'e}-free images for training demoir{\'e}ing networks, the synthesized moir{\'e} patches further undergo an adaptive denoise process to rule out these low-quality moir{\'e} patterns that bear image detail loss. Concretely, we find low-quality pseudo moir{\'e} leads to a large structure difference from its moir{\'e}-free counterpart, which therefore can be removed if the difference score is beyond a threshold adaptive to a particular percentile of the overall structure differences.
Experiments in Table\,\ref{tab:fhdmi} and Table\,\ref{tab:uhdm} demonstrate that, the proposed UnDeM improves the compared baseline by a large margin, on the real moir{\'e} image dataset.
For example, when trained with a size of 384, MBCNN~\cite{zheng2020image} trained on the synthetic images from our UnDeM achieves 19.89 dB in PSNR on FHDMi~\cite{he2020fhde}, while obtaining 19.36 dB from cyclic moir{\'e} learning~\cite{park2022unpaired} and only 9.32 dB from shooting simulation.
Such results not only demonstrate our efficacy, but also enlighten a new moir{\'e} generation method for the demoir{\'e}ing community.

\section{Related Work}
\label{sec:related}

\subsection{Image Demoir{\'e}ing}

Image demoir{\'e}ing target at cleansing moir{\'e} patterns in taken photos. Earlier studies resort to some property presumptions of moir{\'e} patterns, such as space-variant filters~\cite{siddiqui2009hardware,sun2014scanned}, a low-rank constrained sparse matrix decomposition~\cite{liu2015moire,yang2017textured}, and layer decomposition~\cite{yang2017demoireing}.
%
%
Along with the surge of deep learning on many computer vision tasks, demoir{\'e}ing also benefits from the convolutional neural networks (CNNs) recently.
As the pioneering study, Sun \emph{et al}.~\cite{sun2018moire} developed DMCNN, a multi-scale CNN, to remove moir{\'e} patterns at different frequencies and scales.
He \emph{et al}.~\cite{he2019mop} proposed MopNet that is specially designed for unique properties of moir{\'e} patterns including frequencies, colors, and appearances.
Zheng \emph{et al}.~\cite{zheng2020image} introduced a multi-scale bandpass convolutional neural network (MBCNN) that consists of a learnable bandpass filter and a two-step tone mapping strategy to respectively deal with frequency prior and color shift.
Liu \emph{et al}.~\cite{liu2020wavelet} designed WDNet that removes moir{\'e} patterns in the wavelet domain to effectively separate moir{\'e} patterns from image details.
In~\cite{he2020fhde}, a multi-stage framework FHDe$^2$Net is proposed. FHDe$^2$Net employs a global to local cascaded removal branch to erase multi-scale moire patterns and a frequency-based branch to reserve fine details.
Yu \emph{et al}.~\cite{yu2022towards} designed the ESDNet that utilizes the computationally-efficient semantic-aligned scale-aware module to enhance the network’s capability.
However, all these mentioned approaches require large amounts of moir{\'e} and moir{\'e}-free pairs. 
To solve this limitation, a cycle loss is further constructed to simultaneously train a pseudo moir{\'e} generator and a demoir{\'e}ing network~\cite{park2022unpaired,yue2021unsupervised}.
Very differently, our proposed UnDeM does not involve demoire{\'e} network in the moir{\'e} synthesis stage.

\subsection{Moir{\'e}ing Dataset}

Since data-driven CNN-based algorithms require large amount of paired moir{\'e} and moir{\'e}-free images to complete the training,
many efforts have been devoted to constructing large-scale image pairs.
Sun \emph{et al}.~\cite{sun2018moire} built the first real-world moir{\'e} image dataset from ImageNet~\cite{russakovsky2015imagenet}.
He \emph{et al}.~\cite{he2020fhde} proposed the first high-resolution moir{\'e} image dataset FHDMi to satisfy the practical application in the real world. 
Yu \emph{et al}.~\cite{yu2022towards} further proposed the ultra-high-definition demoir{\'e}ing dataset UHDM containing 4K images.
Nevertheless, the data preparation process requires huge human efforts, and the resulting datasets are confined to limited scenes.
To avoid the drudgery of collecting real-world paired moir{\'e} and moir{\'e}-free images, shooting simulation that simulates the camera imaging process becomes a more valuable approach~\cite{liu2018demoir,yuan2019aim}.
However, the synthetic data fails to model the real imaging process and leads to a large domain gap between synthetic data and real data. As a result, demoir{\'e}ing models trained on synthetic data are incapable of handling real-world scenarios.

\begin{figure}[ht]
\centering
\includegraphics[width=0.8\linewidth]{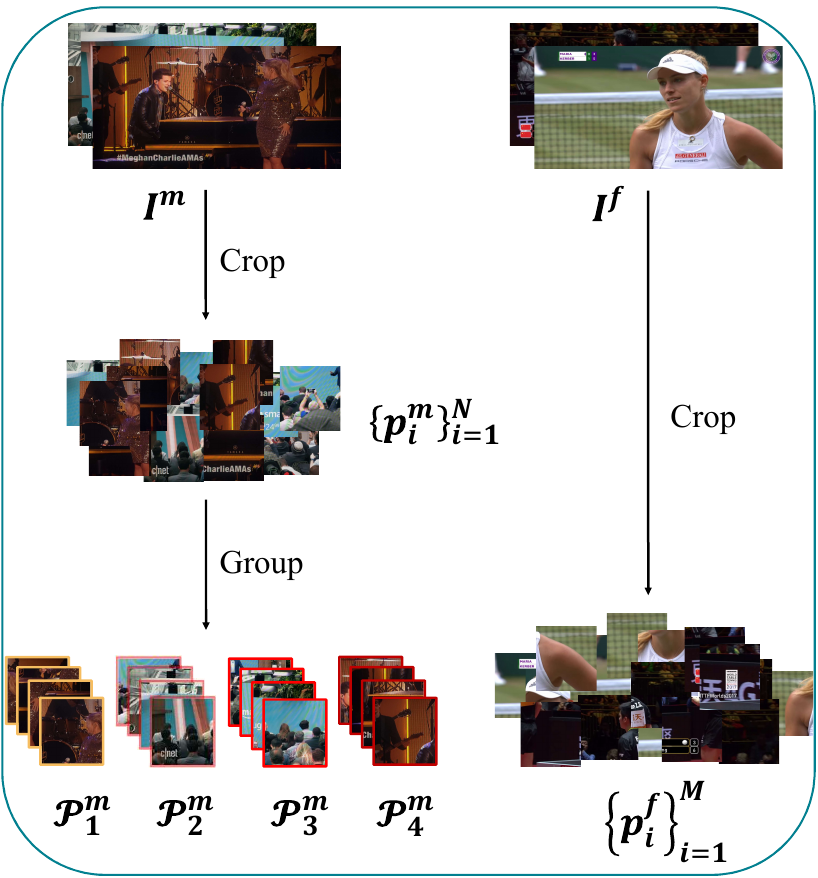}
\caption{Image preprocessing. Both moir{\'e} images $I^m$ and unpaired moir{\'e}-free images $I^f$ are split into patches. Patches from moir{\'e} images are further grouped in compliance with the complexity of moir{\'e} patterns.}
\label{Preprocessing}
\end{figure}

\begin{figure}[ht]
\centering
\includegraphics[width=0.97\linewidth]{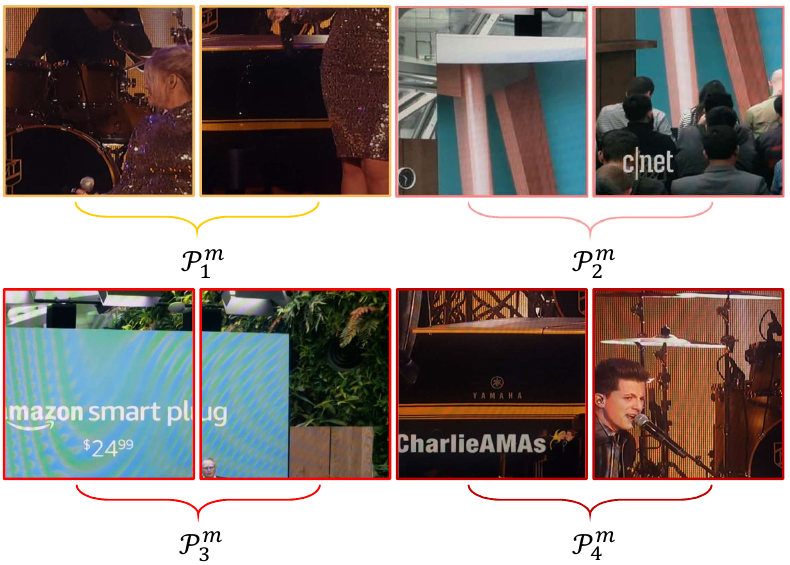}
\caption{An illustration of moir{\'e} images of each group. Each group has its own moir{\'e} patterns complexity.}
\label{group-image}
\end{figure}

\section{Methodology}
Our UnDeM contains image preprocessing, moir{\'e} synthesis network, and adaptive denoise, which are detailed one by one in the following.


\subsection{Image Preprocessing}
\label{sec:prepro}
Moir{\'e} patterns vary significantly even within one single image. 
It is challenging for one single network to learn all cases.
To better learn from these different moir{\'e} patterns, we apply an isolated moir{\'e} synthesis network to deal with moir{\'e} patterns with similar complexity.
We first split images in moir{\'e} set $\mathcal{I}^m$ into non-overlapping patches, leading to a moir{\'e} patch set $\mathcal{P}^m = \{p_i^m\}_{i=1}^N$, where $N$ is the number of patches for the whole moir{\'e} patch set.
Similarly, we can have an $M$-size moir{\'e}-free patch set $\mathcal{P}^f = \{p_i^f\}_{i=1}^M$ for $\mathcal{I}^f$.
As illustrated in Fig.\,\ref{Preprocessing}, we divide the moir{\'e} set $\mathcal{P}^m$ into $K$ subsets $\mathcal{P}^m = \mathcal{P}^m_1 \cup \mathcal{P}^m_2 \cup...\cup \mathcal{P}^m_K$. Each $\mathcal{P}^m_j$ contains moir{\'e} patches with similar complexity and any two subsets are disjoint.
Zhang~\emph{et al}.~\cite{zhang2023real} showed that a perceptible moir{\'e} pattern is highlighted by either high frequency or rich color information. Following~\cite{zhang2023real}, given a moir{\'e} patch $p^m \in \mathcal{P}$, whose frequency is measured by a Laplacian edge detection operator $\mathcal{F}(p^m)$ with kernel size of 3~\cite{marr1980theory}. In addition, the colorfulness, denoted as $\mathcal{C}(p^m)$, is the linear combination of the mean and standard deviation of the pixel cloud in the color plane of RGB colour space~\cite{hasler2003measuring}:
\begin{equation}
\begin{split}
   & \mathcal{C}(p^m) \\& =   \sqrt{ 
            \sigma^2(p^m_R-p^m_G) +\sigma^2\big(0.5(p^m_R+p^m_G)-p^m_B\big)
    } 
    \\&
    + 0.3 \sqrt{ 
            \mu^2(p^m_R-p^m_G) +\mu^2\big(0.5(p^m_R+p^m_G)-p^m_B\big)
    }, 
\end{split}
\end{equation}
where $\sigma(\cdot)$ and $\mu(\cdot)$ return standard deviation and mean value of inputs, the $p^m_R$, $p^m_G$, and $p^m_B$ denote the red, green, and blue color channels of $p^m$.

We set $K = 4$ and obtain four evenly-sized subsets of moir{\'e} patches, each of which has distinctive moir{\'e} features.
The first group $\mathcal{P}_{1}^m$ contains patches with the first $N/4$ smallest $\mathcal{F}(p^m)\cdot\mathcal{C}(p^m)$, thus it has moir{\'e} patterns of low frequency and less color.
We sort the remaining patches from the smallest to the largest with a new metric $\mathcal{F}(p^m)/\mathcal{C}(p^m)$.
Then $\mathcal{P}_2^m$ consists of the first $N/4$ patches highlighted by low frequency but rich color.
The middle $N/4$ patches form $\mathcal{P}_3^m$ featured with high frequency and rich color.
The $N/4$ smallest scored patches with high frequency but less color make up with $\mathcal{P}_4^m$.
Fig.\,\ref{group-image} gives some visual examples.

\begin{figure*}[ht]
\centering
\includegraphics[width=0.78\linewidth]{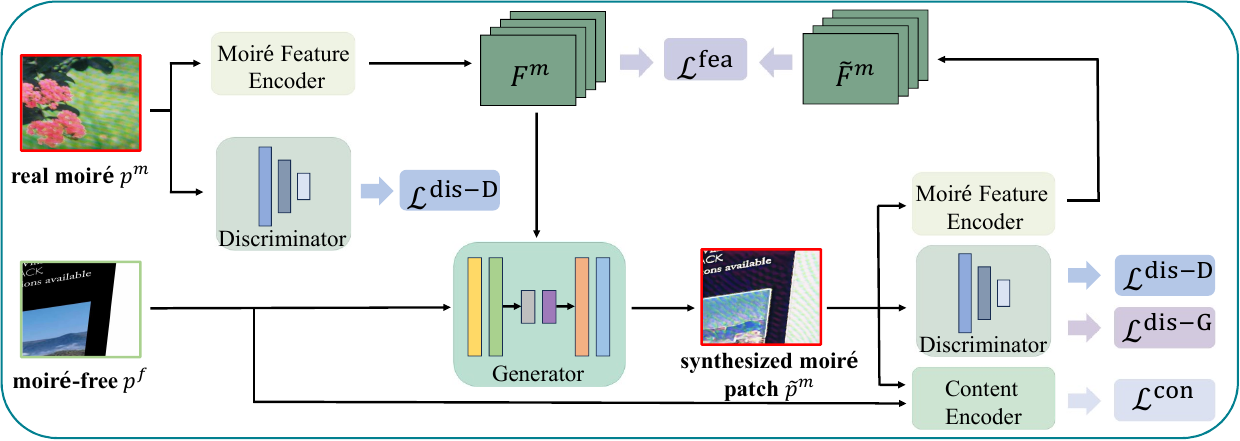}
\caption{Framework of our moir{\'e} synthesis network.}
\label{network}
\end{figure*}

\subsection{Moir{\'e} Synthesis Network}
\label{sec:syn}
Fig.\,\ref{network} depicts an overall framework of our moir{\'e} synthesis network $\mathcal{T}_i$ to learn moir{\'e} patterns from the group $\mathcal{P}_i^m$.
It consists of a moir{\'e} feature encoder $E^m$, a generator $G^m$, a discriminator $D^m$, and a content encoder $E^c$.
Given an unpaired moir{\'e} patch $p^m \in \mathcal{P}^m_i$ and a moir{\'e}-free patch $p^f \in \mathcal{P}^f$, our motivion is to produce a pseudo moir{\'e} $\tilde{p}^m$ that possesses the moir{\'e} pattern of $p^m$ while still retaining image details of $p^f$, such that $(\tilde{p}^m, p^f)$ forms moir{\'e} and moir{\'e}-free pairs to guide the learning of existing demoir{\'e}ing networks.

To fulfill this objective, the moir{\'e} feature encoder $E^m$ extracts the moir{\'e} features of the real moir{\'e} patch $p^m$, denoted as $F^m$:
\begin{equation}
        F^m = E^m(p^m).
\label{ext-Em}
\end{equation}

%
%
%

%
%

%
Then, the generator $G^m$ is to synthesize a pseudo moir{\'e} patch $\tilde{p}^m$ with $F^m$ and $p^f$ as its inputs:
\begin{align}
        \tilde{p}^m = G^m \big( Con(F^m, p^f) \big),
\label{gen-G}
\end{align}
where $Con(\cdot, \cdot)$ indicates the concatenation operation.

%
%
%
%
%
%
%
The discriminator $D^m$ cooperates with the generator $G^m$ for a better pseudo moir{\'e} patch in an adversarial training manner~\cite{goodfellow2014generative}. The generator $G^m$ is trained to trick the discriminator $D^m$ by: 
\begin{equation}
     {\cal L}^\text{dis-G} = \big( D^m(\tilde{p}^m) - 1 \big)^2.
\label{G-Dloss}
\end{equation}

The least squares loss function~\cite{mao2017least} is used for a better training stability. Also, $D^m$ is trained to distinguish the pseudo moir{\'e} patch $\tilde{p}^m$ from the real $p^m$:
\begin{align}
    {\cal L}^\text{dis-D} = \big( D^m(\tilde{p}^m) \big)^2  + \big( D^m(p^m) - 1 \big)^2.
    \label{D-loss}
\end{align}

The loss functions of Eq.\,(\ref{G-Dloss}) and Eq.\,(\ref{D-loss}) are optimized in a min-max game manner. 
As a result, $D^m$ learns to distinguish the pseudo moir{\'e} and the real moir{\'e} images, while the moir{\'e} feature encoder $E^m$ is forced to learn to extract the moir{\'e} feature appropriately and the generator $G^m$ learns to synthesize real-like and in-distribution pseudo moir{\'e} images, 
In addition, we also require moir{\'e} feature of synthesized $\tilde{p}^m$ to follow that of real $p^m$ by:
%
%
%
\begin{align}
    \tilde{F}^m & = E^m(\tilde{p}^m), \\
    {\cal L}^\text{fea} & = \| \tilde{F}^m - F^m  \|_1,
\label{feat-loss}
\end{align}
where $\| \cdot \|_1$ denotes the $\ell_1$ loss. 
To well pair $\tilde{p}^m$ and $p^f$, $\tilde{p}^m$ is also expected to have contents details of $p^f$. An additional content encoder $E^c$ is introduced to align content features between $\tilde{p}^m$ and $p^f$ as:
\begin{equation}
        {\cal L}^\text{con} = \| E^c(\tilde{p}^m) - E^c(p^f)  \|_1.
\label{cont-loss}
\end{equation}

%
%
%
%

%

%
%

Combining Eq.\,(\ref{G-Dloss}), Eq.\,(\ref{D-loss}), Eq.\,(\ref{feat-loss}), and Eq.\,(\ref{cont-loss}) leads to our final loss function as: 
\begin{equation}
    \begin{aligned}
        {\cal L} & = {\cal L}^\text{dis-G}+ {\cal L}^\text{dis-D}  +  {\cal L}^\text{fea} + {\cal L}^\text{con}.
    \end{aligned}
\label{total-loss}
\end{equation}


\begin{figure}[!t]
    \centering
    \begin{subfigure}[b]{\columnwidth}
        \centering
        \includegraphics[width=0.8\textwidth]{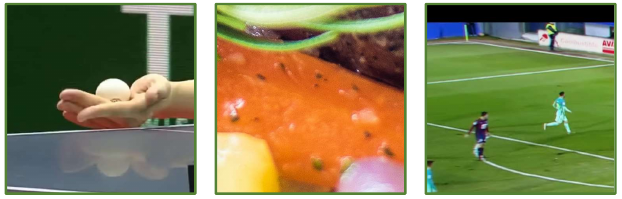} 
        \caption{Moir{\'e}-free patch $p^f$.}
        \label{fig:denoise-clear}
    \end{subfigure}
    \\
    \begin{subfigure}[b]{\columnwidth}
        \centering
        \includegraphics[width=0.8\textwidth]{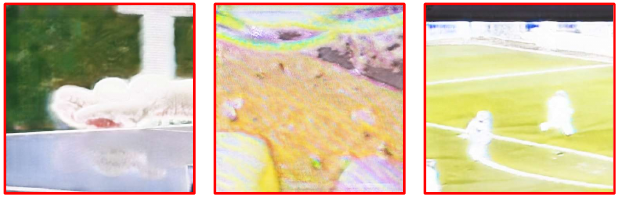} 
        \caption{Low-quality Pseudo moir{\'e} images $\tilde{p}^m$.}
        \label{fig:denoise-ourmo}
    \end{subfigure}
\caption{Examples of low-quality pseudo moir{\'e} images.}
\label{denoise-image}
\end{figure}


\subsection{Adaptive Denoise}
\label{sec:deno}
After training our moir{\'e} synthesis networks $\{\mathcal{T}_i\}_{i=1}^4$, the pseudo moir{\'e} patch $\tilde{p}^m$, paired with corresponding moir{\'e}-free patch $p^f$, sets the dataset to train demoir{\'e}ing network. Unfortunately, we find some pseudo moir{\'e} patches occasionally suffer from low-quality issues. Some examples are manifested in Fig.\,\ref{denoise-image}, where the contents and details of $p^f$ are destroyed in $\tilde{p}^m$. Such noisy data hinders the learning of demoir{\'e}ing models.

Fortunately, we observe in Fig.\,\ref{denoise-image} that the ruined structure mostly attributes to the edge information. Therefore, we calculate the edge map of each patch by the Laplacian edge detection operator, and the structure difference is computed by summing up the absolute value of the edge difference between each pseudo pair.
Low-quality pseudo moir{\'e} leads to a large score of structure difference, and we can rule out these pairs as long as the score is beyond a threshold that is adaptive to the $\gamma$-th percentile of structure differences in a total of $N$ pseudo pairs.
We conduct the above process for each synthesis network $\mathcal{T}_i$ and set a corresponding $\gamma_i$ to remove low-quality pseudo moir{\'e}. We find $N=6,400$ performs well already. Consequently, we obtain a better performance. 

In summary, our UnDeM consists of
1) training a moir{\'e} synthesis network for synthesizing pseudo moir{\'e} images; 2) training a demoir{\'e}ing model based on the trained moir{\'e} synthesis network.
This paper focuses on addressing moir{\'e} image generation. As for demoir{\'e}ing models, we directly borrow from existing studies. Details of training algorithms are listed in the appendix.

\section{Experimention}
\subsection{Implementation Details}
\label{sec:Implementation}

\textbf{Datasets}. 
Public demoir{\'e}ing datasets used in this paper include the FHDMi~\cite{he2020fhde} dataset and UHDM~\cite{yu2022towards} dataset.
The FHDMi dataset consists of 9,981 image pairs for training and 2,019 image pairs for testing with 1920$\times$1080 resolution. 
The UHDM dataset contains 5,000 image pairs with 4K resolution in total, of which 4,500 are used for training and 500 for testing. 
We use the training set to train the proposed moir{\'e} synthesis network.
For image preprocessing, we crop the training images of FHDMi into 8 patches.
For UHDM involving images with higher resolution, we crop the training images into 6 patches. 
During training, the moir{\'e} patch $p^m$ and moir{\'e}-free patch $p^f$ are selected from different original images (before image preprocessing) to ensure they are unpaired.


\textbf{Networks.} 
We implement our UnDeM using the Pytorch framework~\cite{paszke2019pytorch}. The architecture of moir{\'e} synthesis network is largely based on \cite{hu2019mask,liu2021shadowCVPR}.
$E^m$ and $E^c$ contain one convolutional layer and two residual blocks.
$G^m$ contains three convolutional layers, nine residual blocks, and two deconvolutional layers, and ends with a convolutional layer to produce the final output. 
The residual blocks constitute two convolutional layers that are followed by instance normalization and ReLU function~\cite{ulyanov2016instance}.
The convolutional layer has 16 channels for $E^m$ and $E^c$ and 128 for $G^m$. 
$D^m$ is borrowed from PatchGAN~\cite{isola2017image} and consists of three convolutional layers with a stride of 2, two convolutional layers with a stride of 1, and ends with an average pooling layer.
For demoir{\'e}ing models, we utilize MBCNN~\cite{zheng2020image} and ESDNet-L~\cite{yu2022towards} (A large version of ESDNet).

The moir{\'e} synthesis network is trained using the Adam optimizer~\cite{kingma2014adam}, where the first momentum and second momentum are set to 0.9 and 0.999, respectively.
We use 100 epochs for training with a batch size of 4 and an initial learning rate of 2$\times$10$^{-4}$, which is linearly decayed to $0$ in the last 50 epochs.
Besides, we perform different random crop sizes on the image patches after image preprocessing to validate the flexibility of our method for synthesizing pseudo moir{\'e} images.
The crop sizes are set to 192$\times$192 and 384$\times$384 for FHDMi, and 192$\times$192, 384$\times$384, and 768$\times$768 for UHDM, respectively.
As for the demoir{\'e}ing models, we retain the same training configurations as the original paper except that all models are trained for 150 epochs for a fair comparison. 
All networks are initialized using a Gaussian distribution with a mean of 0 and a standard deviation of 0.02. 
The $\gamma_1$, $\gamma_2$, $\gamma_3$, and $\gamma_4$ for adaptive denoise are empirically set to 50, 40, 30, and 20, respectively\footnote{Ablations on $\gamma_i$ and each component in UnDeM are provided in the appendix.}.
All experiments are run on NVIDIA A100 GPUs.

\textbf{Evaluation Protocols}. We adopt the Peak Signal-to-Noise Ratio (PSNR), Structure Similarity (SSIM)~\cite{wang2004image}, and LPIPS~\cite{zhang2018unreasonable} to quantitatively evaluate the performance of demoir{\'e}ing models.

\begin{table}[!t]
\centering
\setlength{\tabcolsep}{0.9mm}{
\begin{tabular}{cccccc}
\toprule[1.25pt]
Model                   & C.S.           & Method & PSNR$\uparrow$ & SSIM$\uparrow$ & LPIPS$\downarrow$ \\ \hline \hline
\multirow{8}{*}{MBCNN}  & \multirow{4}{*}{192} & Paired                   &  22.49    &  0.815  &  0.191   \\
                        &                      & Shooting      &  10.66 &   0.477    & 0.570       \\
                        &                      &  Cyclic     & 19.15  &  0.722      &   0.257     \\
                        &                      & UnDeM                    & 19.45  &  0.732 &  0.230     \\ \cline{3-6} 
                        & \multirow{4}{*}{384} &   Paired        & 22.73  &  0.819 &  0.182    \\
                        &                      & Shooting                & 9.32  &  0.513     &   0.572     \\
                        &                      &    Cyclic       &  19.36 & 0.733   &  0.265     \\
                        &                      &  UnDeM                    & 19.89   & 0.735   &  0.226  \\ \hline
\multirow{8}{*}{ESDNet-L} & \multirow{4}{*}{192} & Paired                    &  22.86 &  0.823    &   0.143     \\
                        &                      &   Shooting        &   10.06   &  0.558     & 0.487        \\
                        &                      &    Cyclic    &  19.09    &   0.738    &   0.241      \\
                        &                      &  UnDeM                    & 19.38 & 0.749   &  0.228     \\ \cline{3-6} 
                        & \multirow{4}{*}{384} & Paired                    &  23.45 &  0.834    & 0.134     \\
                        &                      & Shooting                &   9.81   &  0.553     &  0.512      \\
                        &                      &     Cyclic         &   19.05   & 0.715     &  0.273     \\
                        &                      &  UnDeM                    & 19.66  & 0.747     &   0.205     \\ \bottomrule[0.75pt]
\end{tabular}}
\caption{Quantitative results on the FHDMi dataset. The ``C.S.'' denotes size in the random crop and the ``Paired'' denotes real paired data.
}
\label{tab:fhdmi}
\end{table}

\subsection{Quantitative Results}

\textbf{FHDMi.}
We first analyze the performance on the FHDMi dataset by comparing our UnDeM against the baseline, \emph{i.e.}, the results of the shooting simulation~\cite{niu2021morie} and cyclic learning~\cite{park2022unpaired}.
Table\,\ref{tab:fhdmi} shows that the performance of demoir{\'e}ing models trained on data produced by shooting simulation is extremely poor.
For example, MBCNN obtains only 10.66 dB of PSNR when trained with 192$\times$192 crop size, which indicates the existence of a large domain gap between the pseudo and real data. 
Both the cyclic learning method \cite{park2022unpaired} and our UnDeM exhibit much better results. Moreover, compared with \cite{park2022unpaired}, our UnDeM successfully models the moir{\'e} patterns and thus presents the highest performance.
For instance, MBCNN respectively obtains 19.45 dB and 19.89 dB of PSNR when trained with crop sizes of 192 and 384. 
For ESDNet-L, the PSNR results are 19.38 dB and 19.66 dB, respectively. 
Correspondingly, the SSIM and LPIPS of our UnDeM also exhibit much better performance than shooting simulation and cyclic learning.

\begin{table}[!t]
\centering
\setlength{\tabcolsep}{0.9mm}{
\begin{tabular}{cccccc}
\toprule[1.25pt]
Model                   & C.S.           & Method & PSNR$\uparrow$ & SSIM$\uparrow$ & LPIPS$\downarrow$ \\ \hline \hline
\multirow{12}{*}{MBCNN}  & \multirow{4}{*}{192} & Paired                   & 20.14 &   0.760    &    0.346    \\
                        &                      & Shooting                &  8.99    & 0.528      &  0.632      \\
                        &                      & Cyclic       &  17.42    &  0.663     &  0.464      \\
                        &                      & UnDeM                    & 17.96  & 0.673       &   0.425     \\ \cline{3-6} 
                        & \multirow{4}{*}{384} & Paired                   &  20.14  &  0.759     &   0.356     \\
                        &                      & Shooting                &  9.27   &  0.538     &    0.603    \\
                        &                      & Cyclic     &  17.68    &   0.665    &     0.476   \\
                        &                      &  UnDeM                    &  17.78  &    0.668   &  0.401      \\ \cline{3-6}
                        & \multirow{4}{*}{768} & Paired$\dagger$                   & 21.41 & 0.793      &   0.332    \\
                        &                      & Shooting                &    9.33  &    0.543   &    0.605    \\
                        &                      &  Cyclic  &   17.98   &  0.719     &    0.503    \\
                        &                      &  UnDeM                    &  18.13 &  0.723     & 0.360       \\\hline
\multirow{12}{*}{ESDNet-L} & \multirow{4}{*}{192} & Paired                   & 21.30 & 0.786       &   0.258     \\
                        &                      & Shooting                &  9.80    & 0.606      &   0.544     \\
                        &                      &   Cyclic   &  18.02  &   0.659   &    0.371    \\
                        &                      &  UnDeM                    &   18.30  &  0.662     &  0.365      \\ \cline{3-6} 
                        & \multirow{4}{*}{384} & Paired                   & 21.18  &  0.785     &  0.257      \\
                        &                      & Shooting                &   10.27   &    0.604   &   0.522     \\
                        &                      &  Cyclic    &  17.75    &   0.679    &    0.404    \\
                        &                      &  UnDeM                    &   18.18  &   0.688    &  0.361      \\ \cline{3-6}
                        & \multirow{4}{*}{768} & Paired$\dagger$                   & 22.12 &  0.799    &    0.245   \\
                        &                      & Shooting                &   9.80   &   0.599    &  0.542      \\
                        &                      &   Cyclic  &  18.00  &  0.697     &   0.423     \\
                        &                      &  UnDeM                    &   18.40  &    0.713   & 0.344       \\\bottomrule[0.75pt]
\end{tabular}}
\caption{Quantitative results on the UHDM dataset. The ``C.S.'' denotes size in the random crop and the ``Paired'' denotes real paired data. The ``$\dagger$'' indicates results directly copied from~\cite{yu2022towards}.}
\label{tab:uhdm}
\end{table}

\begin{figure*}[ht]
    \centering
    \begin{subfigure}[b]{\textwidth}
        \centering
        \includegraphics[width=0.85\linewidth]{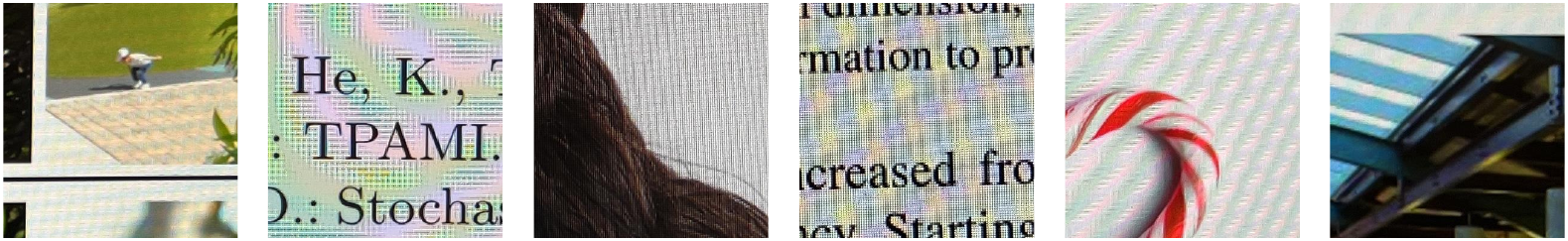} 
        \caption{Moir{\'e} images.}
    \end{subfigure}
    \\
    \begin{subfigure}[b]{\textwidth}
        \centering
        \includegraphics[width=0.85\linewidth]{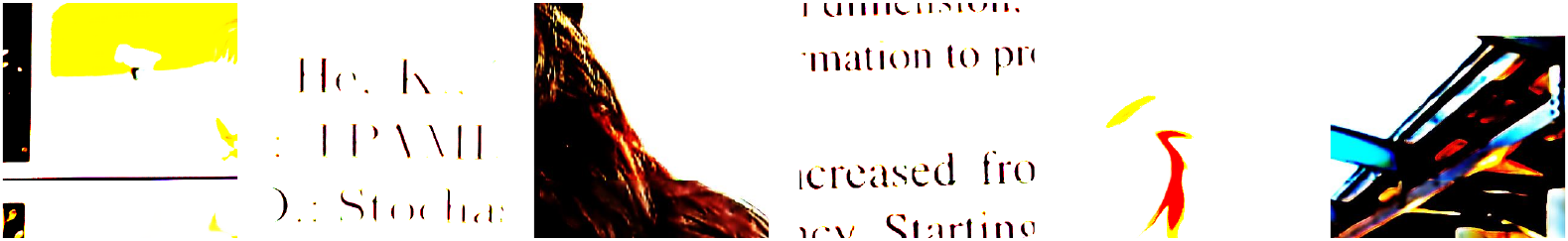} 
        \caption{Demoir{\'e}ing results by shooting simulation~\cite{niu2021morie}.}
        \label{fig:vis-demoire-sim}
    \end{subfigure}
    \\
    \begin{subfigure}[b]{\textwidth}
        \centering
        \includegraphics[width=0.85\linewidth]{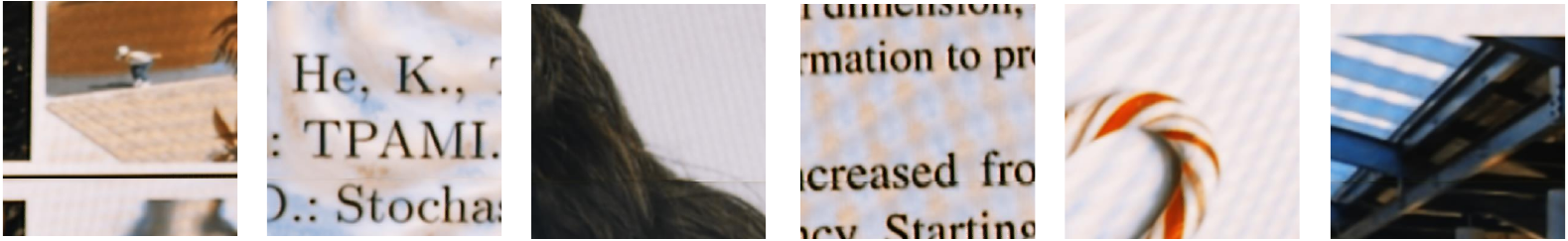} 
        \caption{Demoir{\'e}ing results by cyclic learning~\cite{park2022unpaired}.}
        \label{fig:vis-demoire-baseline}
    \end{subfigure}
    \\
    \begin{subfigure}[b]{\textwidth}
        \centering
        \includegraphics[width=0.85\linewidth]{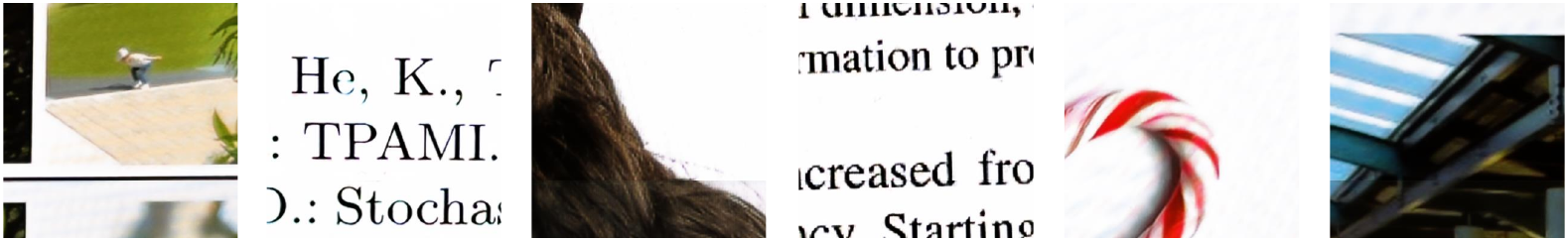} 
        \caption{Demoir{\'e}ing results by our UnDeM.}
        \label{fig:vis-demoire-our}
    \end{subfigure}
    \\
    \begin{subfigure}[b]{\textwidth}
        \centering
        \includegraphics[width=0.85\linewidth]{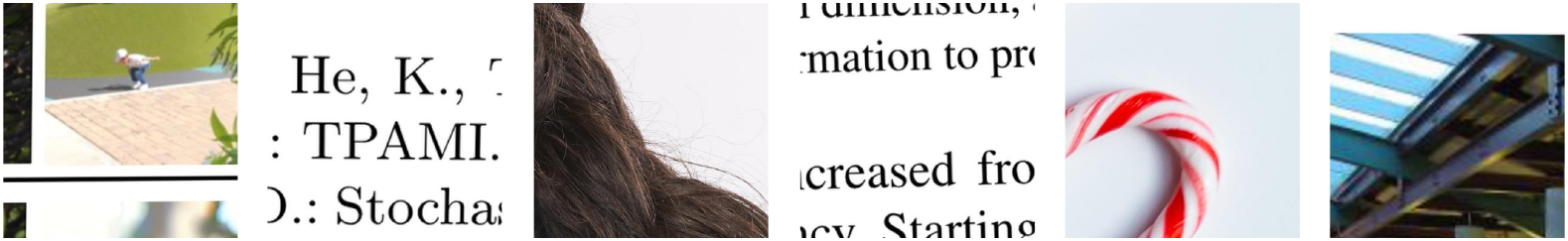} 
        \caption{Moir{\'e}-free images.}
    \end{subfigure}
\caption{Visualization of demoir{\'e}ing results of MBCNN (Crop Size: 768) on the UHDM dataset.  For the convenience of demonstration, we crop the patch from the test image.}
\label{fig:vis}
\end{figure*}


\textbf{UHDM.}
The results on the UHDM dataset are provided in Table\,\ref{tab:uhdm}.
Demoir{\'e}ing models trained on shooting simulation still fail to deal with the real data, and cyclic learning provides better results.
More importantly, our UnDeM surpasses these two methods over different networks and training sizes.
Specifically, the UnDeM increases the PSNR by 0.54 dB, 0.10 dB, and 0.15 dB when training MBCNN with crop sizes of 192, 384, and 768, respectively.
For ESDNet-L, the PSNR gains are 0.28 dB, 0.43 dB, and 0.40, respectively.
To summarize from Table\,\ref{tab:fhdmi} and Table\,\ref{tab:uhdm}, we can conclude that the transferability of our produced moir{\'e} images to the downstream demoir{\'e}ing tasks and the efficacy of our UnDeM over existing methods have therefore been well demonstrated.


\subsection{Qualitative Results}

Qualitative comparisons of demoir{\'e}ing images on UHDM dataset are presented in Fig.\,\ref{fig:vis}, with additional results provided in the appendix.
As shown in Figure \ref{fig:vis-demoire-sim}, the demoir{\'e}ing results of shooting simulation exhibit unnaturally high brightness, leading to a loss of image detail.
This decrease in visual quality can be blamed on the generally darker brightness of shooting simulation as shown in Fig.\,\ref{fig:shooting}, which makes the demoir{\'e}ing model learning incorrect brightness relationship between the moir{\'e} and moir{\'e}-free images.
As presented in Fig.\,\ref{fig:vis-demoire-baseline}, the demoir{\'e}ing model fails to remove moir{\'e} due to cyclic learning cannot model the moir{\'e} patterns as illustrated in Fig.\,\ref{fig:baseline}. Results in Fig.\,\ref{fig:vis-demoire-our} demonstrate the efficacy of UnDeM in removing moir{\'e} patterns, reflecting the fact that UnDeM has the ability to successfully model the moir{\'e} patterns.

\section{Conclusion}

In this paper, we present UnDeM that performs real image demoir{\'e}ing using unpaired real data in a learning-based manner. We synthesize pseudo moir{\'e} images to form paired data for training off-the-shelf demoir{\'e}ing models.
The proposed UnDeM contains three steps including image preprocessing, a moir{\'e} generation network, and adaptive denoise.
The image preprocessing crops the real moir{\'e} images into multiple sub-image patches and groups them into four groups according to the moir{\'e} patterns complexity.
A moir{\'e} generation network is applied to synthesize a pseudo moir{\'e} image that has the moir{\'e} feature as its input real moir{\'e} image and the image detail as its input moir{\'e}-free image.
The adaptive denoise is introduced to rule out the low-quality synthetic moir{\'e} images for avoiding their adverse effects on the learning of demoir{\'e}ing models.
UnDeM is demonstrated to improve the quality of synthetic images and the demoir{\'e}ing models trained on these images are experimentally shown to be superior in performance.

\section{Acknowledgments}

This work was supported by National Key R\&D Program of China (No.2022ZD0118202), the National Science Fund for Distinguished Young Scholars (No.62025603), the National Natural Science Foundation of China (No. U21B2037, No. U22B2051, No. 62176222, No. 62176223, No. 62176226, No. 62072386, No. 62072387, No. 62072389, No. 62002305 and No. 62272401), and the Natural Science Foundation of Fujian Province of China (No.2021J01002,  No.2022J06001).

\bibliography{aaai24}

\clearpage

\appendix

\section{Algorithm Process}

The detailed algorithm process of our method is presented in Alg.\,\ref{supp-alg1}, which consists of 1) training a moir{\'e} synthesis network for synthesizing pseudo moir{\'e} images in Lines $1-6$; 2) training a demoir{\'e}ing model based on the trained moir{\'e} synthesis network in Lines $7-19$.

\begin{algorithm}[!t]
\caption{Overall process.}
\label{supp-alg1} 
    \DontPrintSemicolon
    \LinesNumbered
    \KwIn{Moir{\'e} image set $\mathcal{I}^m$ and moir{\'e}-free image set $\mathcal{I}^f$.}

    \SetKwProg{Fn}{Moir{\'e} synthesis network training}{:}{}
    \Fn{}{
        Split the $\mathcal{I}^m, \mathcal{I}^f$ into $\mathcal{P}^m, \mathcal{P}^f$, and group $\mathcal{P}^m$ into four groups $\{\mathcal{P}^m_i\}_{i=1}^4$ according to their complexity (Sec.3.1 of the main paper).\;
        \For{i = $1$ $\rightarrow$ $4$}{
            Initialize a moir{\'e} synthesis network $\mathcal{T}_i$ and train it with images of $\{\mathcal{P}^m_i\}_{i=1}^4$ (Sec.3.2 of the main paper). \;
            Determine a threshold $\gamma_i$ for adaptive denoise (Sec.3.3 of the main paper). \;
        } 
        \KwRet $\mathcal{P}^f$, $\{\mathcal{P}^m_i\}_{i=1}^4, \{\mathcal{T}_i\}_{i=1}^4, \{\gamma_i\}_{i=1}^4$. \;
    } \;
    \KwIn{$\mathcal{P}^f$, $\{\mathcal{P}^m_i\}_{i=1}^4, \{\mathcal{T}_i\}_{i=1}^4, \{\gamma_i\}_{i=1}^4$.}
    \SetKwProg{Fn}{Demoir{\'e}ing model training}{:}{}
    \Fn{}{
        Initialize a demoir{\'e}ing model $\mathcal{F}$. \;
        \While{not end}{
            Randomly select $\gamma_i$, $\mathcal{P}^m_i$, and $\mathcal{T}_i$. \;
            Randomly select $p^m$ from $\mathcal{P}^m_i$ and $p^f$ from $\mathcal{P}^f$. Use $\mathcal{T}_i$ to synthesize $\tilde{p}^m$ with $p^m$ and $p^f$ as its inputs. \;
            Compute the structure difference $score$ between $\tilde{p}^m$ and $p^f$. \; 
            \eIf{$\gamma_i$ $\leq$ $score$}{
              Update $\mathcal{F}$ with $\tilde{p}^m$ and $p^f$. \;
            }{
              Go to Line 12. \;
            }
        }
        \KwRet $\mathcal{F}$\;
    }

\end{algorithm}

\begin{figure*}[ht]
    \centering
    \begin{subfigure}[b]{0.48\textwidth}
        \centering
        \includegraphics[width=\linewidth]{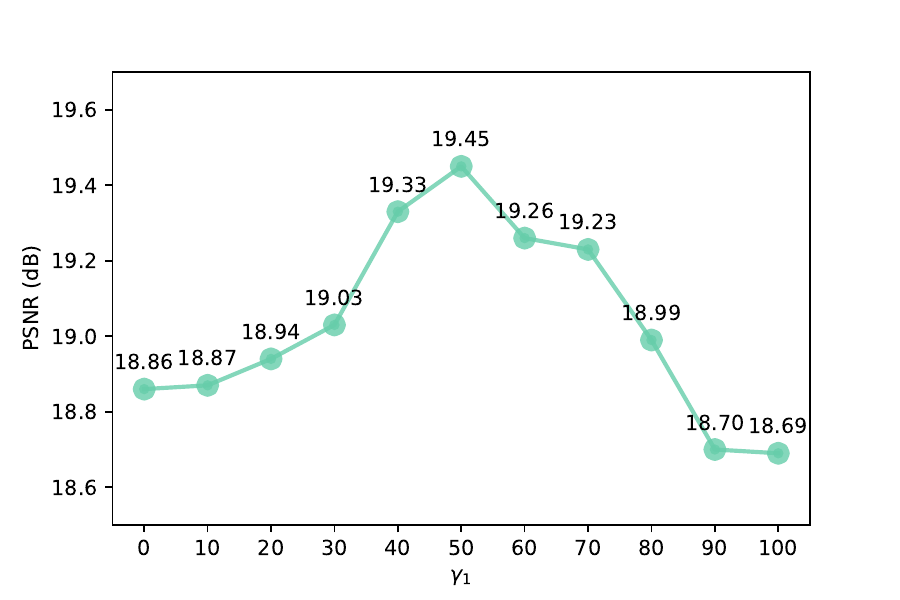} 
        \caption{}
        \label{fig:supp-gamma1}
    \end{subfigure}
    \begin{subfigure}[b]{0.48\textwidth}
        \centering
        \includegraphics[width=\linewidth]{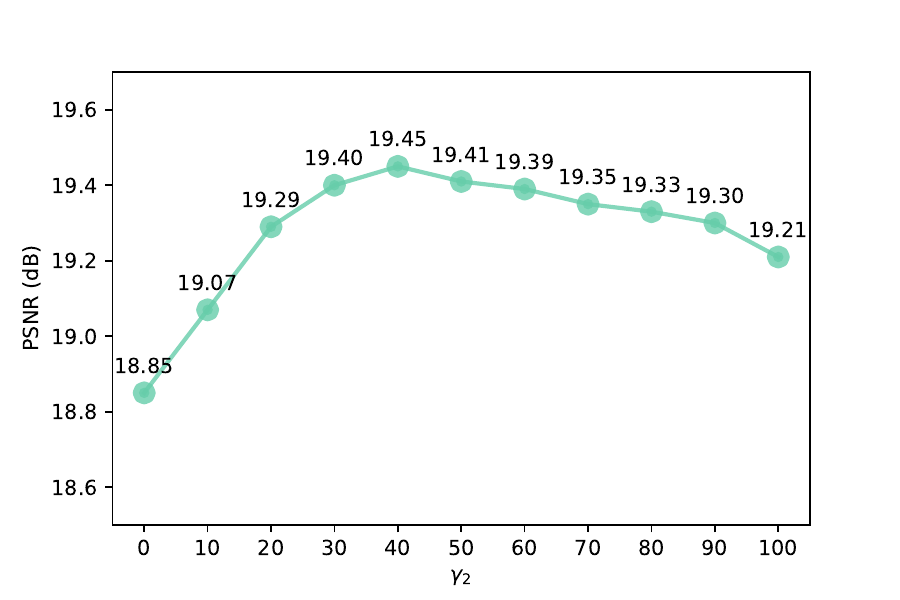} 
        \caption{}
        \label{fig:supp-gamma2}
    \end{subfigure}
    \\
    \begin{subfigure}[b]{0.48\textwidth}
        \centering
        \includegraphics[width=1\linewidth]{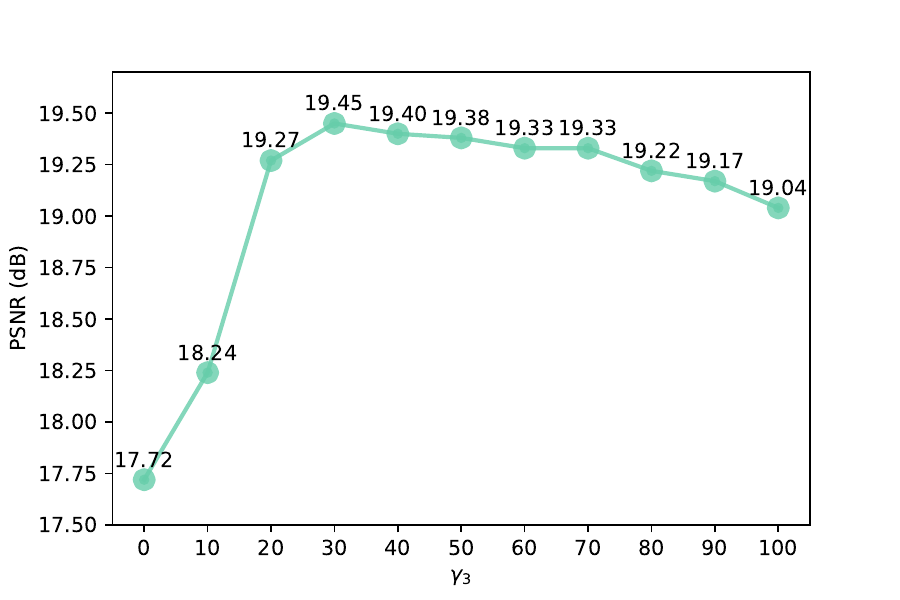} 
        \caption{}
        \label{fig:supp-gamma3}
    \end{subfigure}
    \begin{subfigure}[b]{0.48\textwidth}
        \centering
        \includegraphics[width=1\linewidth]{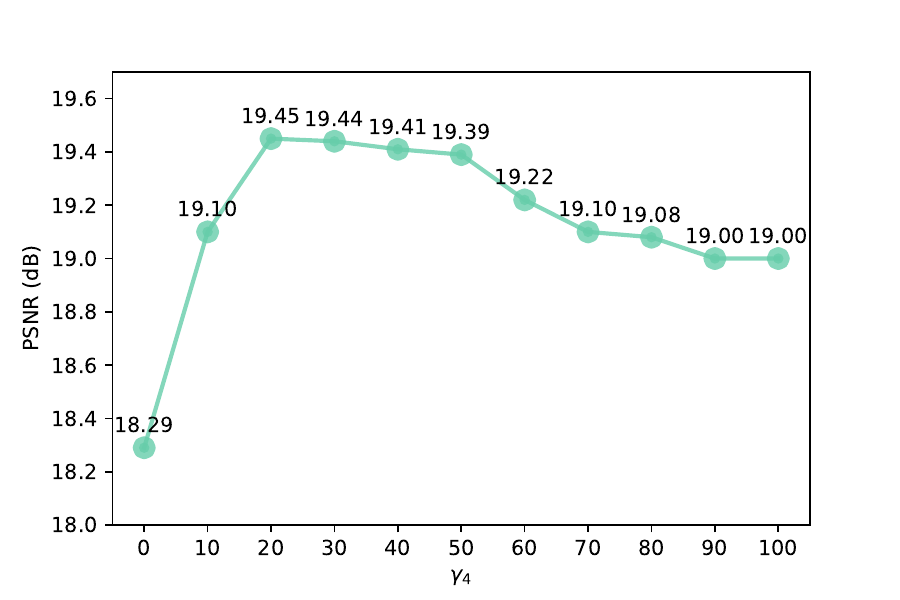} 
        \caption{}
        \label{fig:supp-gamma4}
    \end{subfigure}
\caption{Ablation of $\gamma_i$ in Adaptive Denoise.}
\label{fig:gamma}
\end{figure*}

\section{Ablation of Hyperparameters in Adaptive Denoise}

We present the ablation study of hyperparameters in adaptive denoise, \emph{i.e.}, $\{\gamma_i\}_{i=1}^4$. The experiments are conducted with MBCNN on the FHDMi dataset and the crop size is 192. Each $\gamma_i$ is first empirically initialized. Then, we fix other hyperparameters when performing the ablation study of $\gamma_i$ to search for the optimal value.
From Fig.\,\ref{fig:supp-gamma1} to Fig.\,\ref{fig:supp-gamma4}, we can find that the best configuration for $\gamma_1$, $\gamma_2$, $\gamma_3$, and $\gamma_4$ is 50, 40, 30, and 20, respectively. 
We use these configurations for all crop sizes and networks. Though they might not be optimal under different conditions, they already bring better performance than other methods.

\begin{table}[!t]
\centering
\begin{tabular}{ccccc}
\toprule[1.25pt]
Model                             & Components & PSNR$\uparrow$ & SSIM$\uparrow$ & LPIPS$\downarrow$ \\ \hline \hline
\multirow{6}{*}{MBCNN}  &     All               &  19.45  & 0.732  &  0.230     \\
                   &  w/o Img Pre &  16.62  & 0.645 & 0.309    \\
                   & w/o Denoise  &  18.96   &    0.717 & 0.257 
                   \\
                   &  w/o ${\cal L}^\text{dis}$ &  7.22  & 0.478 & 0.669 \\
                   & w/o ${\cal L}^\text{fea}$  & 18.38  & 0.726 & 0.265   \\
                   & w/o ${\cal L}^\text{con}$  & 17.90     &  0.674 & 0.280         \\\bottomrule[0.75pt]
\end{tabular}
\caption{Ablation study of MBCNN on the FHDMi dataset. The crop size is 192. The ``Img Pre'' denotes image preprocessing and the ``Denoise'' denotes adaptive denoise.}
\label{tab:ablation}
\end{table}

\subsection{Ablation Study}
\label{sec:ablation}

We continue to perform ablation studies to analyze the proposed UnDeM. 
Table\,\ref{tab:ablation} suggests that each component in UnDeM plays an important role in the final demoir{\'e}ing performance. 
In particular, removing image preprocessing and adaptive denoise respectively lead to PSNR drops of 2.83 dB and 0.49 dB.
Also, the performance degrades if removing the losses used in the moir{\'e} synthesis network,~\emph{i.e.}, Eq.\,(4), Eq.\,(5), Eq.\,(7), and Eq.\,(8).
Respectively, removing ${\cal L}^\text{dis}$ (${\cal L}^\text{dis-G}$ and ${\cal L}^\text{dis-D}$), ${\cal L}^\text{fea}$, and ${\cal L}^\text{con}$ causes 12.23 dB, 1.07 dB, and 1.55 dB drop in PSNR.
These results demonstrate the importance of adversarial training for producing real-like pseudo moir{\'e} images.

\begin{figure*}[ht]
    \centering
    \begin{subfigure}[b]{\textwidth}
        \centering
        \includegraphics[width=0.85\linewidth]{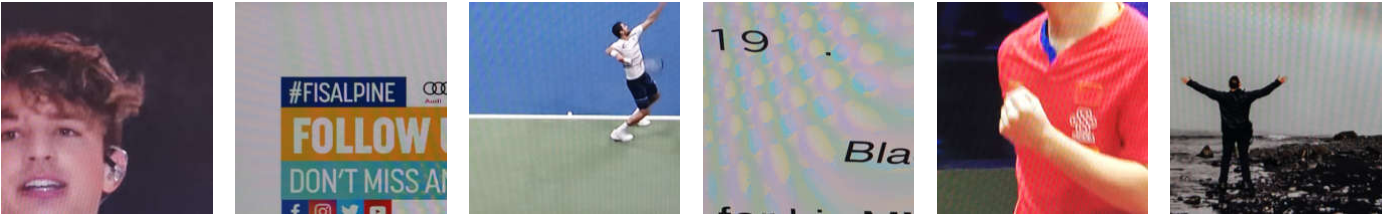} 
        \caption{Moir{\'e} images.}
    \end{subfigure}
    \\
    \begin{subfigure}[b]{\textwidth}
        \centering
        \includegraphics[width=0.85\linewidth]{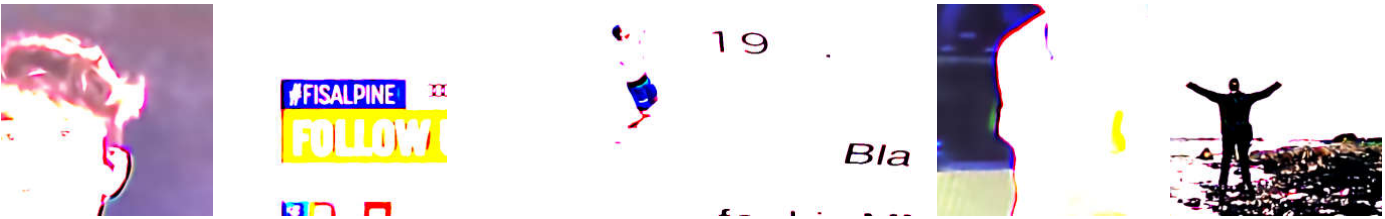} 
        \caption{Demoir{\'e}ing results by shooting simulation~\cite{niu2021morie}.}
    \end{subfigure}
    \\
    \begin{subfigure}[b]{\textwidth}
        \centering
        \includegraphics[width=0.85\linewidth]{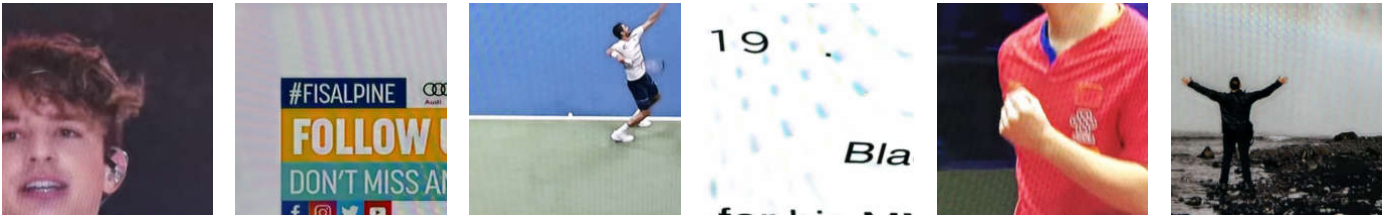} 
        \caption{Demoir{\'e}ing results by cyclic learning~\cite{park2022unpaired}.}
    \end{subfigure}
    \\
    \begin{subfigure}[b]{\textwidth}
        \centering
        \includegraphics[width=0.85\linewidth]{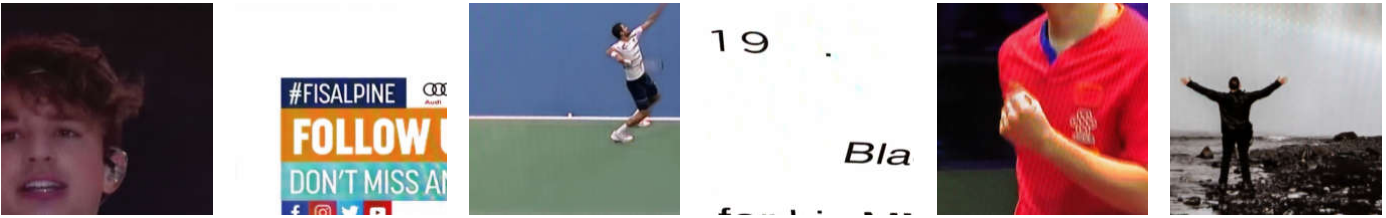} 
        \caption{Demoir{\'e}ing results by our UnDeM.}
    \end{subfigure}
    \\
    \begin{subfigure}[b]{\textwidth}
        \centering
        \includegraphics[width=0.85\linewidth]{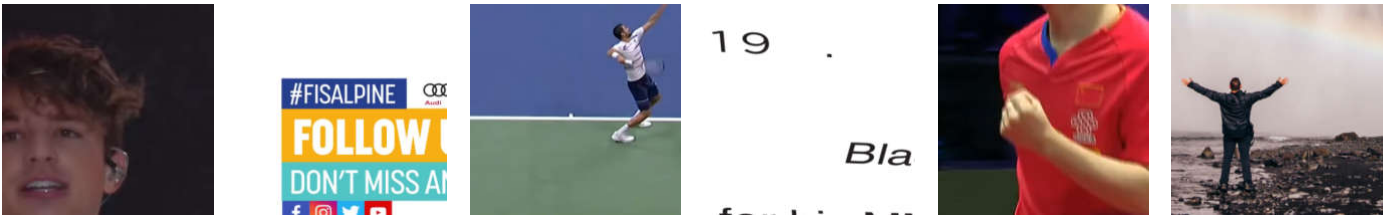} 
        \caption{Moir{\'e}-free images.}
    \end{subfigure}
\caption{Visualization of demoir{\'e}ing results of MBCNN (Crop Size: 384) on the FHDMi dataset. For the convenience of demonstration, we crop the patch from the test image.}
\label{fig:supp-mbcnn-fhdmi}
\end{figure*}

\section{More Qualitative Results}

\begin{figure*}[ht]
    \centering
    \begin{subfigure}[b]{\textwidth}
        \centering
        \includegraphics[width=0.85\linewidth]{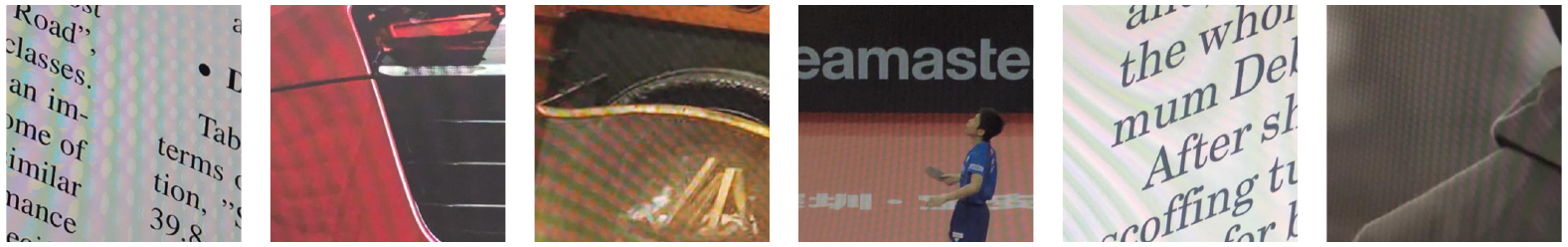} 
        \caption{Moir{\'e} images.}
    \end{subfigure}
    \\
    \begin{subfigure}[b]{\textwidth}
        \centering
        \includegraphics[width=0.85\linewidth]{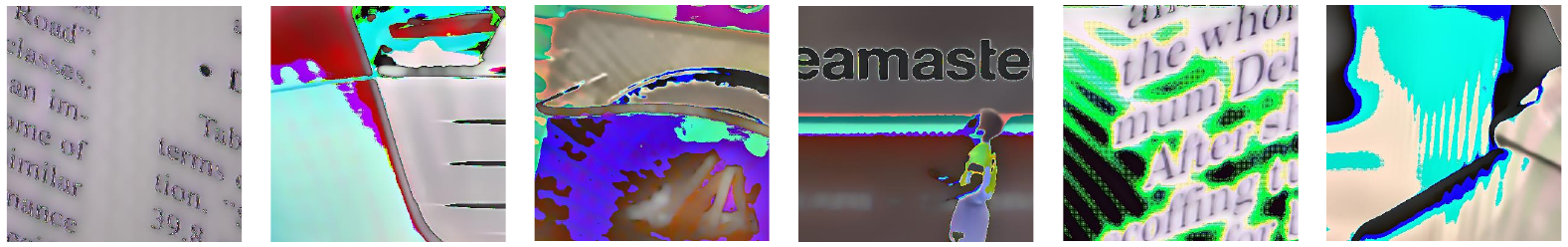} 
        \caption{Demoir{\'e}ing results by shooting simulation~\cite{niu2021morie}.}
    \end{subfigure}
    \\
    \begin{subfigure}[b]{\textwidth}
        \centering
        \includegraphics[width=0.85\linewidth]{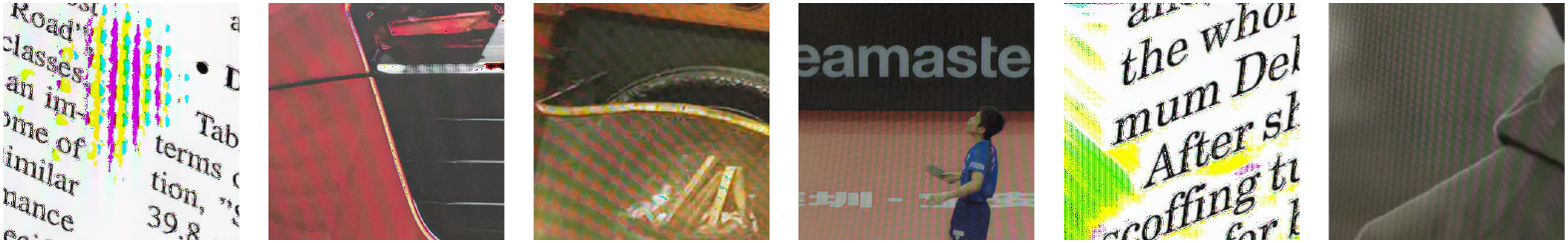} 
        \caption{Demoir{\'e}ing results by cyclic learning~\cite{park2022unpaired}.}
    \end{subfigure}
    \\
    \begin{subfigure}[b]{\textwidth}
        \centering
        \includegraphics[width=0.85\linewidth]{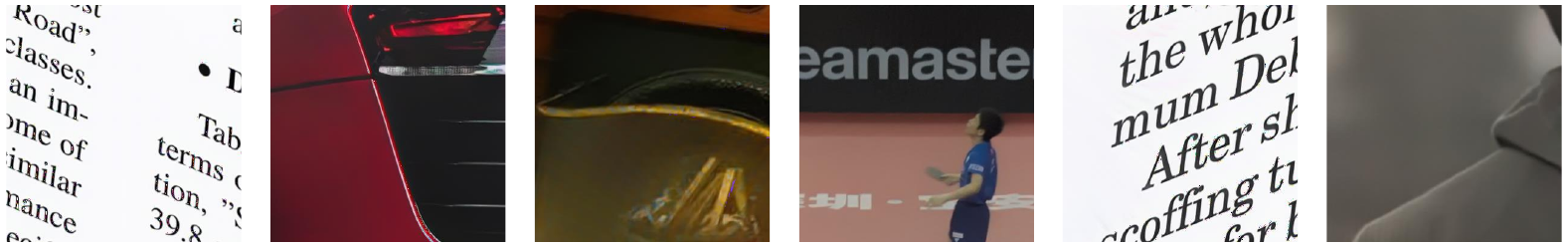} 
        \caption{Demoir{\'e}ing results by our UnDeM.}
    \end{subfigure}
    \\
    \begin{subfigure}[b]{\textwidth}
        \centering
        \includegraphics[width=0.85\linewidth]{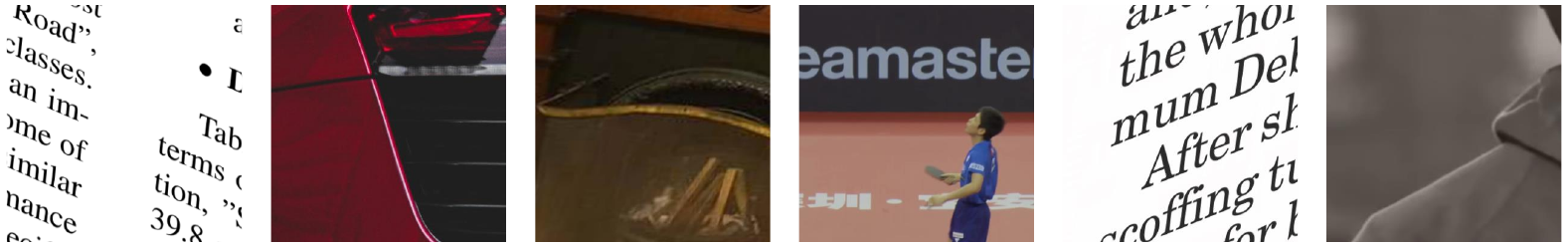} 
        \caption{Moir{\'e}-free images.}
    \end{subfigure}
\caption{Visualization of demoir{\'e}ing results of EDSNet-L (Crop Size: 384) on the FHDMi dataset. For the convenience of demonstration, we crop the patch from the test image.}
\label{fig:supp-esdnet-fhdmi}
\end{figure*}

\begin{figure*}[ht]
    \centering
    \begin{subfigure}[b]{\textwidth}
        \centering
        \includegraphics[width=0.85\linewidth]{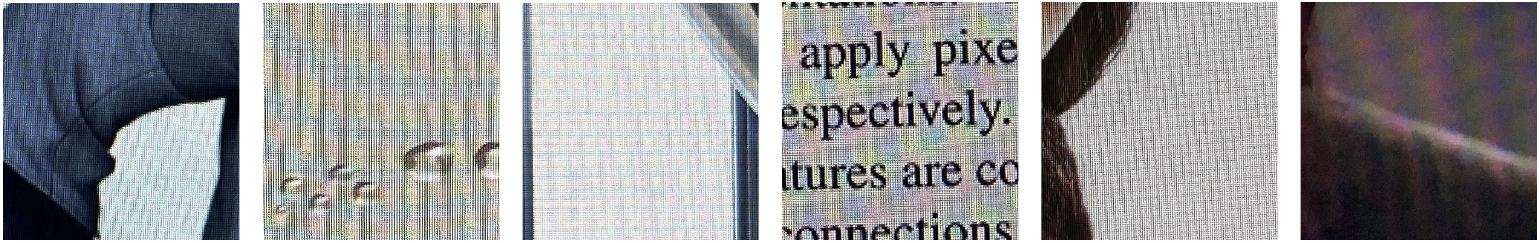} 
        \caption{Moir{\'e} images.}
    \end{subfigure}
    \\
    \begin{subfigure}[b]{\textwidth}
        \centering
        \includegraphics[width=0.85\linewidth]{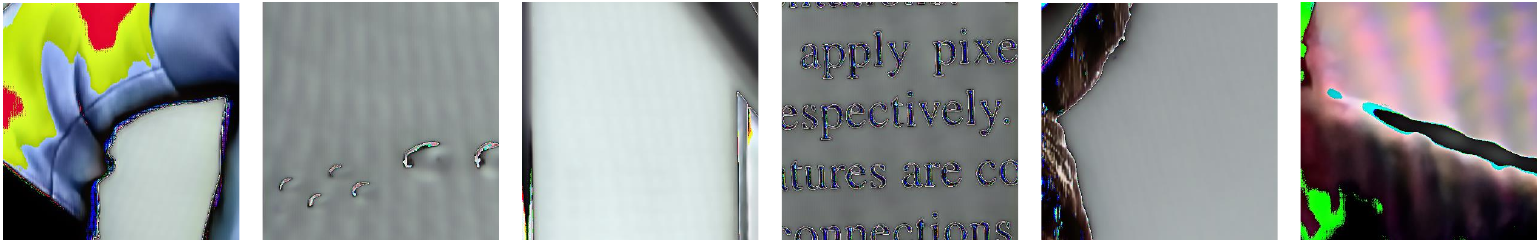} 
        \caption{Demoir{\'e}ing results by shooting simulation~\cite{niu2021morie}.}
    \end{subfigure}
    \\
    \begin{subfigure}[b]{\textwidth}
        \centering
        \includegraphics[width=0.85\linewidth]{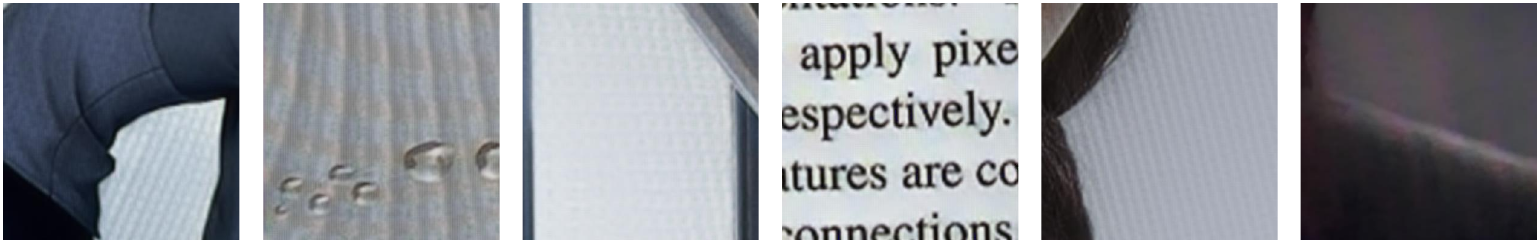} 
        \caption{Demoir{\'e}ing results by cyclic learning~\cite{park2022unpaired}.}
    \end{subfigure}
    \\
    \begin{subfigure}[b]{\textwidth}
        \centering
        \includegraphics[width=0.85\linewidth]{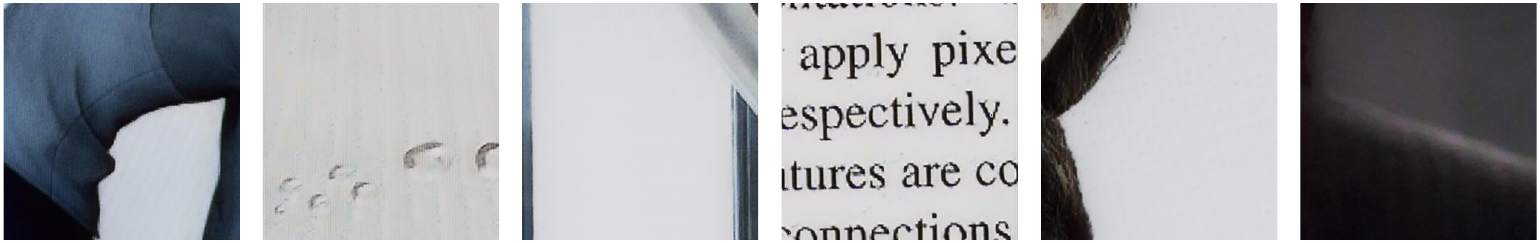} 
        \caption{Demoir{\'e}ing results by our UnDeM.}
    \end{subfigure}
    \\
    \begin{subfigure}[b]{\textwidth}
        \centering
        \includegraphics[width=0.85\linewidth]{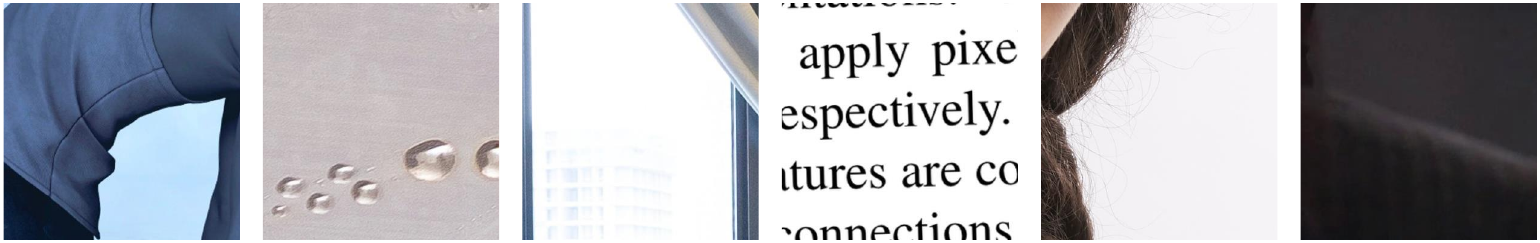} 
        \caption{Moir{\'e}-free images.}
    \end{subfigure}
\caption{Visualization of demoir{\'e}ing results of EDSNet-L (Crop Size: 768) on the UHDM dataset. For the convenience of demonstration, we crop the patch from the test image.}
\label{fig:supp-esdnet-uhdm}
\end{figure*}

In this section, we present more qualitative comparisons. Fig.\,\ref{fig:supp-mbcnn-fhdmi} and Fig.\,\ref{fig:supp-esdnet-fhdmi} present the qualitative results of MBCNN (Crop size: 384) and ESDNet-L (Crop size: 384) on FHDMi dataset, respectively. Fig.\,\ref{fig:supp-esdnet-uhdm} presents the qualitative results of ESDNet-L (Crop size: 768) on UHDM dataset.

\section{Limitations and Discusions}

Although the proposed UnDeM improves the performance of shooting simulation and cyclic learning by a large margin and provides a novel moir{\'e} generation paradigm for the research community, its performance still degrades a lot if compared with the results of real data.
The further improvement of the quality of pseudo moir{\'e} still remains an open question.
We analyze that, a more elaborated image preprocessing, dedicated moir{\'e} synthesis network and special design loss might be potential ways to improve the quality of pseudo images.
Moving forward, further investigation into these potential solutions will be a major focus of our future research efforts.

\end{document}